\documentclass[twoside,leqno,twocolumn]{article}

\usepackage[letterpaper]{geometry}
\usepackage{ltexpprt}
\usepackage{hyperref}
\usepackage{tablefootnote}
\usepackage{dirtytalk}
\usepackage{xcolor}
\usepackage{soul}
\usepackage{times}
\usepackage{latexsym}
\usepackage{cleveref}
\usepackage{url}

\usepackage{graphics}
\usepackage{graphicx}
\usepackage{comment}
\usepackage{caption}
\usepackage{appendix}

\usepackage{multirow}
\usepackage[para]{threeparttable}

\title{\Large A Roadmap for  Multilingual, Multimodal Domain Independent Deception Detection\thanks{All authors are with the University of Houston}}

\author{Dainis Boumber \\
  {\tt dboumber@uh.edu}
\and Rakesh M. Verma \\
  {\tt rmverma2@Central.UH.EDU}
\and Fatima Zahra Qachfar \\
  {\tt fqachfar@uh.edu}
}
\date{}

\date{}

\begin{document}
\maketitle
\begin{abstract}

Deception, a prevalent aspect of human communication, has undergone a significant transformation in the digital age. With the globalization of online interactions, individuals are communicating in multiple languages and mixing languages on social media, with varied data becoming available in each language and dialect. At the same time, the techniques for detecting deception are similar across the board. Recent studies have shown the possibility of the existence of universal linguistic cues to deception across domains within the English language; however, the existence of such cues in other languages remains unknown. Furthermore, the practical task of deception detection in low-resource languages is not a well-studied problem due to the lack of labeled data. Another dimension of deception is multimodality. For example, a picture with an altered caption in fake news or disinformation may exist. This paper calls for a comprehensive investigation into the complexities of deceptive language across linguistic boundaries and modalities within the realm of computer security and natural language processing and the possibility of using multilingual transformer models and labeled data in various languages to universally address the task of deception detection.

\end{abstract}

\section{Introduction}

Deception is a complex and pervasive phenomenon with profound implications for various domains, including security, law enforcement, healthcare, and human-computer interaction. Accurately identifying deception has long been a critical goal for researchers and practitioners alike. Traditional methods for deception detection (DD) have primarily relied on linguistic cues and textual analysis~\cite{verma2022domainindependent,10.1145/3508398.3519358}. A DD task is typically a binary classification problem, aiming to label a statement as being deceptive or not. Less often, the goal is to categorize a statement as falling into one of the more or less deceptive categories. It is a problem of growing importance that is made more challenging by the need to build different datasets and detectors for the ever-increasing variety of domains and tasks where deceptive language poses a threat. However, these methods often fall short in the face of sophisticated deceivers who can manipulate language effectively, leaving the task of deception detection far from foolproof~\cite{mehdi2023adversarial}. Recently, there has been a paradigm shift towards more comprehensive and robust approaches to deception detection, which leverage multimodal data sources. This shift recognizes that deception is not confined to language alone and that individuals may convey deceptive information through various channels, including speech, facial expressions, body language, and by using different languages\footnote{However,  multilingual deception detection efforts are relatively few.} Another frequently debated topic is the transfer of linguistic cues of deception across domains and modalities. The need for domain-independence in deception detection is paramount since there are many manifestations of deception. 

We call for a holistic approach to deception detection, focusing on integrating multimodal (multiple modes of communication) and multilingual (cross-linguistic) data while maintaining domain independence. We propose leveraging cutting-edge advances in natural language processing (NLP), computer vision, and machine learning (ML) to enhance the accuracy and robustness of deception detection across a wide array of applications and settings. The research in this area must address several critical challenges in deception detection, including integrating non-verbal cues from multiple modalities (e.g., speech, facial expressions, gestures, image/video attachments) and considering linguistic variations across different languages. By developing a domain-independent approach, we call for creating a versatile approach that can be applied to diverse scenarios, from border security and criminal investigations to healthcare diagnostics and online content moderation.

This paper will present the theoretical foundations of multimodal multilingual deception detection through existing work and suggest the methodologies to be employed. We aim to open a new direction of research that will usher valuable insights into a comprehensive approach to deception detection that transcends linguistic and contextual boundaries, opening up new possibilities for enhancing trust, security, and decision-making across various domains.

\section{The Idea and Impetus}

History is replete with famous lies and deceptions. Examples include P. T. Barnum, Nicolo Machiavelli, Sun Tzu, Operation Mincemeat, and the Trojan Horse~\cite{encyclopedia_deception}. A chronology of deception is included in~\cite{encyclopedia_deception}. More recently, 
the proliferation of deceptive attacks such as fake news, phishing, and disinformation is rapidly eroding trust in Internet-dependent societies. The situation has deteriorated so much that 45\% of the US population believes the 2020 US election was stolen.\footnote{\url{https://www.surveymonkey.com/curiosity/axios-january-6-revisited}.}
 
Social media platforms have come under severe scrutiny regarding how they police content. Facebook and Google partner with independent fact-checking organizations that typically employ manual fact-checkers. Things will only get worse with the advent of Large Language Models, such as ChatGPT. 

Natural-language processing (NLP) and machine learning (ML) researchers have joined the fight by designing fake news, phishing, and other types of domain-specific detectors. Building single-domain detectors may be suboptimal. Composing them sequentially requires more time, and composing them in parallel requires more hardware. Moreover, building single-domain detectors means one can only react to new forms of deception {\bf after}  they emerge. We aim to stimulate research on \emph{ domain-independent} deception. Unfortunately, research in this area is currently hampered by the lack of computational definitions and taxonomy, high-quality datasets, and systematic approaches to domain-independent deception detection. Thus, the results are neither generalizable nor reliable, leading to much confusion. \emph{By domain independence, we mean that deception can have many different goals and motivations, such as phishing, job scams, political lies, fake news, etc., not just different topical content} For example, in previous work, researchers used ``multi-domain'' to refer to lies on abortion, the death penalty, or feelings about best friends (see related work in ~\cite{capuozzo2020decop}). 

Below, we briefly survey some of the work that has been done on deception so far. Of course, there is a lot of work on phishing, fake news, etc., when considered in isolation. Still, there are hardly any works on identifying common patterns in different deceptive attacks and some have even claimed that there are no common linguistic cues of deception~\cite{grondahl2019text}.

\section{Related Work}

In the past few years, there have been several studies of applying computational methods to deal with deception detection in a single domain. For fake news, {\cite{ceron2020fake}} used topic models and {\cite{hamid2020fake}} used  Bag of Words (BoW) and BERT {\cite{devlin2019bert}} embedding. The state of the art (SOTA) in phishing detection has been dominated by classical supervised machine learning approaches and deep neural nets~\cite{el2020depth}. More recently, BERT~\cite{devlin2019bert}, a character-level CNN, and sentence embeddings from Sentence-BERT (SBERT)~\cite{reimers2019sentencebert} were used to find emails exhibiting psychological traits most dominant in phishing texts~\cite{shahriar2022improving}. In detecting opinion spam and fake reviews, weakly supervised graph networks have been recently used with some success~\cite{Li2022ShootingRS}. {\cite{Feng2012SyntacticSF,Mukherjee2013WhatYF,inproceedings}} used part-of-speech tags and context-free grammar parse trees, behavioral features, and spatial-temporal features, respectively. Neural network methods for spam detection consider the reviews as input without specific feature extraction. In {\cite{article}}, authors used a gated recurrent neural network to study the contextual information of review sentences. DRI-RCNN {\cite{Zhang2018DRIRCNNAA}} used a recurrent network for learning the contextual information of the words in the reviews. Several studies on cross-domain deception detection have been published, as well {\cite{HernndezCastaeda2017CrossdomainDD,RillGarca2018FromTT,SnchezJunquera2020MaskingDI}}. Recently, a quality domain-independent deception dataset was introduced in~{\cite{10.1145/3508398.3519358}}, with the empirical evidence suggesting that large language models such as BERT and RoBERTa perform well on individual tasks when fine-tuned on a combination of out of domain deceptive texts. 
Closer to our stated goals,~\cite{Volkova2019ExplainingMD} created a multi-modal deception detection tool that used early deep learning models and word embeddings, although ultimately, the performance was not always robust and it lacked domain-independence capabilities. Finally,
\cite{glenski-etal-2020-towards} propose a framework for evaluating the robustness of deception detection models across two domains (Twitter and Reddit), modalities (Text, images), and five languages.

The authors in \cite{glenski-etal-2020-towards} propose a framework to evaluate the robustness of deception detection models in two domains (Twitter and Reddit), modalities (text, images), and five languages (English, French, German, Russian,
and Spanish) highlighting similar challenges. In our topic, we would like to extend the investigation further to include domain-independent deception instead of domain-specific models as discussed in \cite{glenski-etal-2020-towards}.

\paragraph{Datasets}
We recommend utilizing a range of deception data sources to create a diverse and versatile dataset for training AI models in multilingual and multimodal deception detection. A good starting point is provided in \Cref{tab:data_sources}, where we have emphasized multi-domain datastes, but this list is by no means complete or exhaustive,\footnote{Especially since there is a plethora of datasets for single domains such as fake news or disinformation.} and finding relevant data is not a finite process; that is, more data should be added as new datasets are found. This approach will help ensure that models are not limited to a single language, or a single modality, or a single domain, and can effectively detect deception across different cultures, communication channels, and domains.

\begin{table*}[!ht]
\footnotesize
\centering
\caption{Deception Detection Dataset Sources. Multi-domain datasets are in boldface font in the Description column.}
\label{tab:data_sources}
\begin{threeparttable}
\begin{tabular}{|c|c|c|l|c|c|c|}
\hline
\textbf{Language} & \textbf{Description} & \textbf{\begin{tabular}[c]{@{}c@{}}Contains \\ Synthetic \\ Data\end{tabular}} &  \multicolumn{1}{c|}{\textbf{Labels}} & \textbf{\begin{tabular}[c]{@{}c@{}}Total\\ Samples\end{tabular}} & \textbf{Distribution} & \textbf{Source}  
\\ \hline

\multirow{6}{*}{\textbf{Arabic}}  & Hate Speech & No & \begin{tabular}[l]{@{}l@{}}Toxic \\ Not Toxic\end{tabular} & 1k & Balanced & SurgeAI \tnote{1} 

\\ \cline{2-7} 

& \multirow{4}{*}{{\bf Disinformation Detection} \cite{araieval:arabicnlp2023-overview}} & \multirow{4}{*}{No} & \begin{tabular}[l]{@{}l@{}}disinfo \\ no-disinfo\end{tabular} & $>$ 19k & \multirow{4}{*}{Imbalanced}  & \multirow{4}{*}{\begin{tabular}[c]{@{}c@{}}ArAIEval \\ Gitlab \end{tabular}} \tnote{2} 
\\ \cline{4-5}

& & & \begin{tabular}[l]{@{}l@{}}Rumor, Spam \\ Offensive (OFF), \\ Hate Speech (HS)\end{tabular}  & 3.9k &  & 
\\ \cline{2-7} 
& ArCOV19-Rumors \cite{haouari2020arcov19} & No & \begin{tabular}[l]{@{}l@{}}false, true, other\end{tabular}  & 9.4k
 & Imbalanced & \begin{tabular}[c]{@{}c@{}} bigIR \\ Gitlab\end{tabular} \tnote{3} 

 \\ \hline
\hline

\multirow{6}{*}{\textbf{English}} & {\bf GDD} \cite{10.1145/3508398.3519358} & No & false, true & $>$ 50k & Imbalanced & Zenodo \tnote{4} 

 \\ \cline{2-7} 
& Facebook Misinformation & No & Misinformation & 529 & One label & SurgeAI \tnote{5} 

 \\ \cline{2-7} 
& Yelp Reviews & No & \begin{tabular}[l]{@{}l@{}} Genuine (1) \\ Fake (-1)\end{tabular} & 35.9k & Imbalanced & Kaggle \tnote{6}  

\\ \cline{2-7} 

& Amazon Fake Reviews \cite{hussain2020spam} & No & \begin{tabular}[l]{@{}l@{}}Spam (1) \\ Not Spam (0)\end{tabular} & 26.7M & Imbalanced & Kaggle \tnote{7}  

\\ \cline{2-7} 

& Generated Fake Reviews \cite{salminen2022creating} & Yes & \begin{tabular}[l]{@{}l@{}} generated \\ real \end{tabular} & 40k & Balanced &  OSF\tnote{8} 

\\ \hline
\hline

\textbf{Japanese} & \begin{tabular}[c]{@{}c@{}} {\bf Hate Speech}, \\ {\bf Insults, and Toxicity}\end{tabular} & No & \begin{tabular}[l]{@{}l@{}}Toxic \\ Not Toxic\end{tabular} & 1k & Balanced & SurgeAI \tnote{9} 

\\ \hline
\hline
\textbf{Spanish} & Hate Speech & No & \begin{tabular}[l]{@{}l@{}}Toxic \\ Not Toxic\end{tabular} & 1k & Balanced & SurgeAI \tnote{10} 
\\ \hline
\hline
\textbf{Greek} & elAprilFoolsCorpus~\cite{Papantoniouetal2021elaprilfool} & No & \begin{tabular}[l]{@{}l@{}}Deceptive \\ Truthful\end{tabular} & 508 & Balanced & Gitlab \tnote{11} 

\\ \hline
\textbf{Multilingual} & Fake News~\cite{li2020mm} & No &  \begin{tabular}[l]{@{}l@{}}Fake News \\ Truthful\end{tabular} & $>$ 11k & Imbalanced & Gitlab\tnote{12}
\\ \hline
\end{tabular}
\end{threeparttable}
\begin{tablenotes}
\item[1] \url{https://www.surgehq.ai/datasets/arabic-hate-speech-dataset}

\item[2] \url{https://gitlab.com/araieval/wanlp2023_araieval/-/tree/main/task2}

\item[3] \url{https://gitlab.com/bigirqu/ArCOV-19/-/tree/master/ArCOV19-Rumors}

\item[4] \url{https://zenodo.org/record/6512468}

\item[5] \url{https://www.surgehq.ai/datasets/facebook-misinformation-dataset}

\item[6] \url{https://www.kaggle.com/datasets/abidmeeraj/yelp-labelled-dataset}

\item[7] \url{https://www.kaggle.com/datasets/naveedhn/amazon-product-review-spam-and-non-spam}

\item[8] \url{https://osf.io/tyue9/\#!}

\item[9] \url{https://app.surgehq.ai/datasets/japanese-toxicity}

\item[10] \url{https://www.surgehq.ai/datasets/spanish-hate-speech-dataset}

\item[11] \url{https://gitlab.isl.ics.forth.gr/papanton/elaprilfoolcorpus}

\item[12] \url{https://github.com/bigheiniu/MM-COVID}
\end{tablenotes}
\end{table*}


\section{Challenges and Future Opportunities}

In this section, we briefly discuss significant research challenges and point out future research opportunities. One of the main problems that makes this an open-ended idea is the fact that there is no consensus on the transferability of deceptive cues across domains, even within a single modality. For example, a recent review of deception literature~\cite{grondahl2019text} found unclear and contradictory results and concluded there was no evidence of deception’s stylistic trace. On the other hand, a more recent publication by~\cite{10.1145/3508398.3519358} found evidence to the contrary insofar as DD within the text. Other substantial challenges include:

\begin{enumerate}
    \item \textit{Defining deception computationally.} So far, deception has been defined using the intent of the deceiver, but the attacker is elusive in the real world, so intentions are impossible to access. We point the reader to~\cite{verma2024domain} for a new definition and taxonomy. 
    \item Giving a taxonomy for deception that is comprehensive and useful to guide further research. For example, the taxonomy should help build a quality general deception dataset and then generalized deception detection models. Making sure it is high quality can also be a challenge (but see~\cite{verma2019data,dang2024data}).
    \item Finding a common basis for the different forms of deception.  
    \item Finding common linguistic cues and invariants across the different forms of deception. Some evidence is reported in~\cite{verma2024domain}.
    \item \textit{Dealing with imbalanced data.} Deceptive attacks, by their nature, would be targeted, e.g., spearphishing, or overly broad, such as spam or phishing. This gives rise to imbalanced scenarios. An adapter, Prexia, is reported in~\cite{lrec2024}.
    
    \item \textit{Distributed nature.} People and companies are uncomfortable sharing sensitive information, such as targeted attacks (spearphishing). Can we design models that can work with limited shared data?
    
    \item \textit{Human in the loop.} Can the detector improve human ability? Can humans improve the detector's ability with just a few examples? Or by providing access to his/her cognitive load through a sensor?
\end{enumerate} 

\section{Defining Success}

Ideally, we would achieve a deception detection model that does not need any labeled data to detect new attacks since it is based on invariants of deception. However, this may be too difficult a holy grail to achieve. Thus, success would be a detector that helps the human in the loop do significantly better at resisting attacks (e.g., a novice email user can detect quite sophisticated spearphishing attacks). If we can achieve this goal, then we can start researching the problem of building teachable detectors so that the human and the detector can improve each other. 

\section{Possible Solution}

In this section, we present a preliminary solution to address the problem outlined in this work. Our proposed approach leverages the power of advanced, large-scale, multilingual, and multimodal contextual learners, complemented by Retrieval Augmented Generation (RAG). It is important to note that this approach is just one of several promising avenues that warrant further exploration.

To lay the foundation for a potential solution, we initially focus on a single modality, namely text. In this configuration, we aim to create a system capable of identifying deceptive text in a domain-agnostic and multilingual context. To ensure the system's relevance and effectiveness, we aim to imbue it with desirable attributes, including result explainability and robust zero to few-shot performance.
To achieve these characteristics, a logical system design might involve the integration of an exceptionally large-scale language model that excels in contextual learning, such as Mistral~\cite{jiang2023mistral}. This model can achieve superior performance in the task, regardless of the domain, and remains resilient in the face of diverse data distributions. To further enhance its explainability, the model can undergo instruction tuning, enabling it to deliver answers and elucidate its underlying reasoning.

To elevate its already outstanding zero to few-shot capabilities and achieve performance parity with fully fine-tuned specialized models in specific domains, it is advisable to augment the learner with a Retrieval Augmented Generation (RAG) infrastructure, as proposed by~\cite{lewis2021retrievalaugmented}. This design decision ensures the longevity of the system and its' ability to stay up-to-date with current threats because it makes continuous finetuning unnecessary.

\begin{figure}[!ht]
\begin{center}
\includegraphics[width=1\columnwidth]{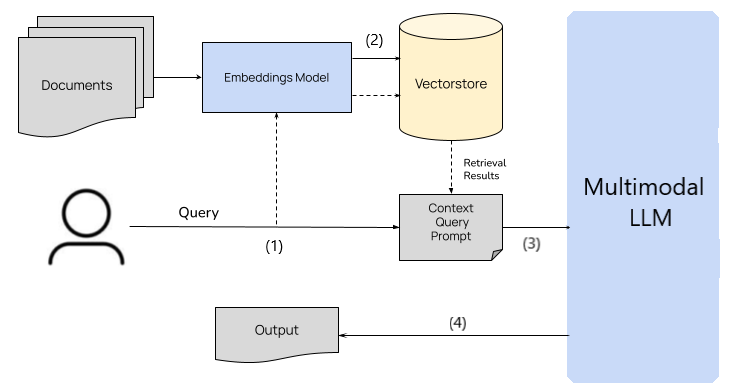} 
\caption{High-level overview of a possible solution using RAG and a multi-modal in-context learner. The dashed line depicts the retrieval of context and its integration into the query.}
\label{fig:rag}
\end{center}
\end{figure}

A standard RAG system's high-level system design scheme is shown in Figure \ref{fig:rag}. It consists of mining-derived datasets indexed and stored in a vector database, accessed through FAISS \cite{douze2024faiss} (Facebook AI Similarity Search) or any other approximate nearest neighbors algorithm optimized for searching large vector spaces. Combined with Retrieval-Augmented Generation (RAG) techniques \cite{lewis2021retrievalaugmented}, this infrastructure delivers critical context to enhance Mistral's performance. This system design is a decision-making tool that detects deceptive text followed by a clear explanation of this decision. Moreover, enhancing the LLM input prompt with retrieved context guarantees the model has all the necessary information to generate a comprehensive response.

When presented with a single instance of a task, typically a query from a user, the steps in solving the problem (each step is referenced in Figure \ref{fig:rag} as (1),(2),(3), or (4)) are:
\begin{enumerate}
    \item \textit{Create Initial Prompt:} Starting with the user query.
    \item \textit{Augment Prompt with Retrieved Context:} Merges the initial prompt with the context retrieved from the Vector Store, creating an enriched input for the LLM.
    \item  \textit{Send Augmented Prompt to LLM:} The LLM receives the enhanced prompt.
    \item \textit{Receive LLM’s Response:} After processing the augmented prompt, it generates its response.
\end{enumerate}

A multimodal LLM is necessary to extend this approach to multiple modalities, e.g., a unified multimodal model, or UnIVAL \cite{shukor2023unival}, which unifies text, images, video, and audio into a single model. Thus, it may be possible to use UnIVAL, or other multimodal models such as OPENAI's CLIP\footnote{\url{https://www.pinecone.io/learn/series/image-search/clip/}}, in place of a single-mode model like Mistral \cite{jiang2023mistral}. In addition, other components, such as RAG, may need to be adjusted as needed.  However, most problems to be solved are in the engineering space.

From a research perspective, the main challenge in designing and implementing a functional system as described is model performance as a function of model size and the current lack of multimodal in-context learners that are large enough to perform satisfactorily. For text, we notice that models of 2-3B parameters score 0.25 or so on the HuggingFace LLM benchmark, 7B score 0.73, and 170B  score 0.75; in other words, there is a massive jump from 3B to 7B  parameters followed by a plateau. So, to have a truly intelligent model, it needs to be at least 7B for a single modality, as a rule of thumb - although it may be possible to bring this number down through quantization and other means. Multiple modalities may require larger models for similar performance. Currently, UnIVAL has only 0.25B parameters.
Therefore, we hypothesize that such a system as we described would be made possible by an advance in models like UnIVAL - multimodal transformers that learn in context and have billions of parameters. 
Perhaps, Flamingo with 80B parameters~\cite{alayrac2022flamingo} can help here.

\section{Conclusions}

In this paper, we introduce a new concept \say{Multilingual, Multimodal Domain-independence  Deception Detection} that unifies diverse investigations, creating a new paradigm for detecting deceitful behavior across languages and modalities. This innovative approach harmoniously connects previous research in multimodal and cross-lingual deception detection, paving the way for future breakthroughs. We discuss the research challenges related to this concept and also potential solutions.

\section*{Acknowledgments.} Research partly supported by NSF grants 2210198 and 2244279, and ARO grants W911NF-20-1-0254 and W911NF-23-1-0191. Verma is the founder of Everest Cyber Security and Analytics, Inc. 

\bibliographystyle{siam}
\bibliography{siam_references}

\begin{thebibliography}{10}

\bibitem{alayrac2022flamingo}
{\sc J.-B. Alayrac, J.~Donahue, P.~Luc, A.~Miech, I.~Barr, Y.~Hasson, K.~Lenc, A.~Mensch, K.~Millican, M.~Reynolds, R.~Ring, E.~Rutherford, S.~Cabi, T.~Han, Z.~Gong, S.~Samangooei, M.~Monteiro, J.~Menick, S.~Borgeaud, A.~Brock, A.~Nematzadeh, S.~Sharifzadeh, M.~Binkowski, R.~Barreira, O.~Vinyals, A.~Zisserman, and K.~Simonyan}, {\em Flamingo: a visual language model for few-shot learning}, 2022.

\bibitem{lrec2024}
{\sc D.~Boumber, F.~Z. Qachfar, and R.~M. Verma}, {\em Domain-agnostic adapter architecture for deception detection: Extensive evaluations with the difraud benchmark}, in Joint International Conference on Computational Linguistics, Language Resources and Evaluation, Torino, Italy, May 2024, European Language Resources Association.

\bibitem{capuozzo2020decop}
{\sc P.~Capuozzo, I.~Lauriola, C.~Strapparava, F.~Aiolli, and G.~Sartori}, {\em Decop: A multilingual and multi-domain corpus for detecting deception in typed text}, in Proceedings of the Twelfth Language Resources and Evaluation Conference, 2020, pp.~1423--1430.

\bibitem{ceron2020fake}
{\sc W.~Ceron, M.-F. de~Lima-Santos, and M.~G. Quiles}, {\em Fake news agenda in the era of covid-19: Identifying trends through fact-checking content. online social networks and media, 21, 100116}, 2020.

\bibitem{dang2024data}
{\sc V.~M.~H. Dang and R.~M. Verma}, {\em Data quality in {NLP}: Metrics and a comprehensive taxonomy}, in International Symposium on Intelligent Data Analysis, Springer, 2024, pp.~217--229.

\bibitem{devlin2019bert}
{\sc J.~Devlin, M.-W. Chang, K.~Lee, and K.~Toutanova}, {\em Bert: Pre-training of deep bidirectional transformers for language understanding}, 2019.

\bibitem{douze2024faiss}
{\sc M.~Douze, A.~Guzhva, C.~Deng, J.~Johnson, G.~Szilvasy, P.-E. Mazaré, M.~Lomeli, L.~Hosseini, and H.~Jégou}, {\em The faiss library}, arXiv preprint arXiv:2401.08281,  (2024).

\bibitem{el2020depth}
{\sc A.~El~Aassal, S.~Baki, A.~Das, and R.~M. Verma}, {\em An in-depth benchmarking and evaluation of phishing detection research for security needs}, IEEE Access, 8 (2020), pp.~22170--22192.

\bibitem{Feng2012SyntacticSF}
{\sc S.~Feng, R.~Banerjee, and Y.~Choi}, {\em Syntactic stylometry for deception detection}, in Annual Meeting of the Association for Computational Linguistics, 2012.

\bibitem{glenski-etal-2020-towards}
{\sc M.~Glenski, E.~Ayton, R.~Cosbey, D.~Arendt, and S.~Volkova}, {\em Towards trustworthy deception detection: Benchmarking model robustness across domains, modalities, and languages}, in Proceedings of the 3rd International Workshop on Rumours and Deception in Social Media (RDSM), Barcelona, Spain (Online), Dec. 2020, Association for Computational Linguistics, pp.~1--13.

\bibitem{grondahl2019text}
{\sc T.~Gr{\"o}ndahl and N.~Asokan}, {\em Text analysis in adversarial settings: Does deception leave a stylistic trace?}, ACM Computing Surveys (CSUR), 52 (2019), pp.~1--36.

\bibitem{hamid2020fake}
{\sc A.~Hamid, N.~Shiekh, N.~Said, K.~Ahmad, A.~Gul, L.~Hassan, and A.~Al-Fuqaha}, {\em Fake news detection in social media using graph neural networks and nlp techniques: A covid-19 use-case}, arXiv preprint arXiv:2012.07517,  (2020).

\bibitem{haouari2020arcov19}
{\sc F.~Haouari, M.~Hasanain, R.~Suwaileh, and T.~Elsayed}, {\em Arcov19-rumors: Arabic covid-19 twitter dataset for misinformation detection}, arXiv preprint arXiv:2010.08768,  (2020).

\bibitem{araieval:arabicnlp2023-overview}
{\sc M.~Hasanain, F.~Alam, H.~Mubarak, S.~Abdaljalil, W.~Zaghouani, P.~Nakov, G.~Da~San~Martino, and A.~Freihat}, {\em Araieval shared task: Persuasion techniques and disinformation detection in arabic text}, in Proceedings of the First Arabic Natural Language Processing Conference (ArabicNLP 2023), Singapore, Dec. 2023, Association for Computational Linguistics.

\bibitem{HernndezCastaeda2017CrossdomainDD}
{\sc {\'A}.~Hern{\'a}ndez-Casta{\~n}eda, H.~Calvo, A.~Gelbukh, and J.~J.~G. Flores}, {\em Cross-domain deception detection using support vector networks}, Soft Computing, 21 (2017), pp.~585--595.

\bibitem{hussain2020spam}
{\sc N.~Hussain, H.~T. Mirza, I.~Hussain, F.~Iqbal, and I.~Memon}, {\em Spam review detection using the linguistic and spammer behavioral methods}, IEEE Access, 8 (2020), pp.~53801--53816.

\bibitem{jiang2023mistral}
{\sc A.~Q. Jiang, A.~Sablayrolles, A.~Mensch, C.~Bamford, D.~S. Chaplot, D.~de~las Casas, F.~Bressand, G.~Lengyel, G.~Lample, L.~Saulnier, L.~R. Lavaud, M.-A. Lachaux, P.~Stock, T.~L. Scao, T.~Lavril, T.~Wang, T.~Lacroix, and W.~E. Sayed}, {\em Mistral 7b}, 2023.

\bibitem{encyclopedia_deception}
{\sc T.~R. Levine}, {\em Encyclopedia of Deception}, vol.~2, Sage Publications, 2014.

\bibitem{lewis2021retrievalaugmented}
{\sc P.~Lewis, E.~Perez, A.~Piktus, F.~Petroni, V.~Karpukhin, N.~Goyal, H.~Küttler, M.~Lewis, W.~tau Yih, T.~Rocktäschel, S.~Riedel, and D.~Kiela}, {\em Retrieval-augmented generation for knowledge-intensive nlp tasks}, 2021.

\bibitem{inproceedings}
{\sc H.~Li, Z.~Chen, A.~Mukherjee, B.~Liu, and J.~Shao}, {\em Analyzing and detecting opinion spam on a large-scale dataset via temporal and spatial patterns}, in Proceedings of the international AAAI conference on web and social media, vol.~9, 2015, pp.~634--637.

\bibitem{Li2022ShootingRS}
{\sc J.~Li, L.~Yang, and P.~Zhang}, {\em Shooting review spam with a weakly supervised approach and a sentiment-distribution-oriented method}, Applied Intelligence, 53 (2022), pp.~10789--10799.

\bibitem{li2020mm}
{\sc Y.~Li, B.~Jiang, K.~Shu, and H.~Liu}, {\em Mm-covid: A multilingual and multimodal data repository for combating covid-19 disinformation}, arXiv preprint arXiv:2011.04088,  (2020).

\bibitem{mehdi2023adversarial}
{\sc P.~Mehdi~Gholampour and R.~M. Verma}, {\em Adversarial robustness of phishing email detection models}, in Proceedings of the 9th ACM International Workshop on Security and Privacy Analytics, 2023, pp.~67--76.

\bibitem{Mukherjee2013WhatYF}
{\sc A.~Mukherjee, V.~Venkataraman, B.~Liu, and N.~S. Glance}, {\em What yelp fake review filter might be doing?}, Proceedings of the International AAAI Conference on Web and Social Media,  (2013).

\bibitem{Papantoniouetal2021elaprilfool}
{\sc K.~Papantoniou, P.~Papadakos, G.~Flouris, and D.~Plexousakis}, {\em Linguistic cues of deception in a multilingual april fools' day context}, in Proceedings of the Eighth Italian Conference on Computational Linguistics, CLiC-it 2021, Milan, Italy, January 26-28, 2022, E.~Fersini, M.~Passarotti, and V.~Patti, eds., vol.~3033 of {CEUR} Workshop Proceedings, CEUR-WS.org, 2021.

\bibitem{reimers2019sentencebert}
{\sc N.~Reimers and I.~Gurevych}, {\em Sentence-bert: Sentence embeddings using siamese bert-networks}, 2019.

\bibitem{article}
{\sc Y.~Ren and D.~Ji}, {\em Neural networks for deceptive opinion spam detection: An empirical study}, Information Sciences, 385-386 (2017), pp.~213--224.

\bibitem{RillGarca2018FromTT}
{\sc R.~Rill-Garc{\'i}a, L.~Villase{\~n}or-Pineda, V.~Reyes-Meza, and H.~J. Escalante}, {\em From text to speech: A multimodal cross-domain approach for deception detection}, in CVAUI/IWCF/MIPPSNA@ICPR, 2018.

\bibitem{salminen2022creating}
{\sc J.~Salminen, C.~Kandpal, A.~M. Kamel, S.-g. Jung, and B.~J. Jansen}, {\em Creating and detecting fake reviews of online products}, Journal of Retailing and Consumer Services, 64 (2022), p.~102771.

\bibitem{SnchezJunquera2020MaskingDI}
{\sc J.~S{\'a}nchez-Junquera, L.~Villase{\~n}or-Pineda, M.~M. y~G{\'o}mez, P.~Rosso, and E.~Stamatatos}, {\em Masking domain-specific information for cross-domain deception detection}, Pattern Recognit. Lett., 135 (2020), pp.~122--130.

\bibitem{shahriar2022improving}
{\sc S.~Shahriar, A.~Mukherjee, and O.~Gnawali}, {\em Improving phishing detection via psychological trait scoring}, 2022.

\bibitem{shukor2023unival}
{\sc M.~Shukor, C.~Dancette, A.~Rame, and M.~Cord}, {\em Unival: Unified model for image, video, audio and language tasks}, Transactions on Machine Learning Research Journal,  (2023).

\bibitem{verma2024domain}
{\sc R.~M. Verma, N.~Dershowitz, V.~Zeng, D.~Boumber, and X.~Liu}, {\em Domain-independent deception: A new taxonomy and linguistic analysis}, arXiv preprint arXiv:2402.01019,  (2024).

\bibitem{verma2022domainindependent}
{\sc R.~M. Verma, N.~Dershowitz, V.~Zeng, and X.~Liu}, {\em Domain-independent deception: Definition, taxonomy and the linguistic cues debate}, 2022.

\bibitem{verma2019data}
{\sc R.~M. Verma, V.~Zeng, and H.~Faridi}, {\em Data quality for security challenges: Case studies of phishing, malware and intrusion detection datasets}, in Proceedings of the 2019 ACM SIGSAC Conference on Computer and Communications Security, 2019, pp.~2605--2607.

\bibitem{Volkova2019ExplainingMD}
{\sc S.~Volkova, E.~Ayton, D.~L. Arendt, Z.~Huang, and B.~Hutchinson}, {\em Explaining multimodal deceptive news prediction models}, in International Conference on Web and Social Media, 2019.

\bibitem{10.1145/3508398.3519358}
{\sc V.~Zeng, X.~Liu, and R.~M. Verma}, {\em Does deception leave a content independent stylistic trace?}, in Proceedings of the Twelfth ACM Conference on Data and Application Security and Privacy, CODASPY '22, New York, NY, USA, 2022, Association for Computing Machinery, p.~349–351.

\bibitem{Zhang2018DRIRCNNAA}
{\sc W.~Zhang, Y.~Du, T.~Yoshida, and Q.~Wang}, {\em Dri-rcnn: An approach to deceptive review identification using recurrent convolutional neural network}, Inf. Process. Manag., 54 (2018), pp.~576--592.

\end{thebibliography}

\end{document}